\documentclass{article}

\usepackage{amsmath}
\usepackage{amsfonts}
\usepackage{bm}

\def\eqref#1{equation~\ref{#1}}

\def\1{\bm{1}}

\DeclareMathAlphabet{\mathsfit}{\encodingdefault}{\sfdefault}{m}{sl}
\SetMathAlphabet{\mathsfit}{bold}{\encodingdefault}{\sfdefault}{bx}{n}

\newcommand{\cod}{\textsc{CoDBench}}

\usepackage{hyperref}
\usepackage{url}
\usepackage[numbers]{natbib}
\usepackage{float}
\usepackage{tikz}
\usepackage{forest}
\usepackage{graphicx}
\usetikzlibrary{trees}
\usepackage{colortbl}
\usepackage{xcolor}
\usepackage{tabularx}
\usepackage[normalem]{ulem}
\useunder{\uline}{\ul}{}
\usepackage{booktabs}
\usepackage{placeins}
\usepackage[para]{footmisc}
\usepackage{paracol}
\usepackage[preprint]{neurips_2023}
\usepackage[utf8]{inputenc} 
\usepackage[T1]{fontenc}    
\usepackage{hyperref}       
\usepackage{url}            
\usepackage{booktabs}       
\usepackage{amsfonts}       
\usepackage{nicefrac}       
\usepackage{microtype}      
\usepackage{xcolor}         
\usepackage{subfig}

\title{\cod: A Critical Evaluation of Data-driven Models for Continuous Dynamical Systems}

\newcommand{\fnn}{\textsc{FNN}}
\newcommand{\resnet}{\textsc{ResNet}}
\newcommand{\unet}{\textsc{UNet}}
\newcommand{\cgan}{\textsc{cGAN}}
\newcommand{\fno}{\textsc{FNO}}
\newcommand{\deeponet}{\textsc{DeepONet}}
\newcommand{\poddeeponet}{\textsc{POD-DeepONet}}
\newcommand{\wno}{\textsc{WNO}}
\newcommand{\sno}{\textsc{SNO}}
\newcommand{\oformer}{\textsc{OFormer}}
\newcommand{\gnot}{\textsc{GNOT}}

\newcommand{\burgers}{\textsc{Burgers}}
\newcommand{\darcy}{\textsc{Darcy}}
\newcommand{\navierstokes}{\textsc{Navier Stokes}}
\newcommand{\shallowwater}{\textsc{Shallow Water}}
\newcommand{\stress}{\textsc{Stress}}
\newcommand{\strain}{\textsc{Strain}}
\newcommand{\shear}{\textsc{Shear}}
\newcommand{\biaxial}{\textsc{Biaxial}}
\newcommand{\csa}{\textsc{Computer Science and Automation}}
\newcommand{\ece}{\textsc{Electrical Communication Engineering}}

\begin{document}

\footnotetext[1]{\csa}
\footnotetext[2]{\ece\newline}
\footnotetext[3]{Upon acceptance, all codes and datasets utilized in this study will be made publicly accessible through GitHub and available for future exploration.}
\author{%
  Priyanshu Burark\\
  \footnotemark[1]{}\hspace{0.18cm}Department of CSA\\
  Indian Institute of Science, Bangalore\\
  Bengaluru, 560012, India \\
  \texttt{priyanshub@iisc.ac.in} \\
  \And
  Karn Tiwari,  Prathosh A P\\
  \footnotemark[2]{}\hspace{0.18cm}Department of ECE\\
  Indian Institute of Science, Bangalore\\
  Bengaluru, 560012, India \\
  \texttt{\{karntiwari, prathosh\}@iisc.ac.in} \\
  \And
  Meer Mehran Rashid\\
  Department of Civil Engineering\\
  Indian Institute of Technology, Delhi\\
  New Delhi, 110016, India\\
  \texttt{mrashi12@jhu.edu} \\
  \And
  N M Anoop Krishnan\\
  Yardi School of Artificial Intelligence\\
  Indian Institute of Technology, Delhi\\
  New Delhi, 110016, India \\
  \texttt{krishnan@iitd.ac.in} \\
}

\maketitle

\begin{abstract}
  Continuous dynamical systems, characterized by differential equations, are ubiquitously used to model several important problems: plasma dynamics, flow through porous media, weather forecasting, and epidemic dynamics. Recently, a wide range of data-driven models has been used successfully to model these systems. However, in contrast to established fields like computer vision, limited studies are available analyzing the strengths and potential applications of different classes of these models that could steer decision-making in scientific machine learning. Here, we introduce \cod, an exhaustive benchmarking suite comprising 11 state-of-the-art data-driven models for solving differential equations. Specifically, we comprehensively evaluate 4 distinct categories of models, \textit{viz.}, feed forward neural networks, deep operator regression models, frequency-based neural operators, and transformer architectures against 8 widely applicable benchmark datasets encompassing challenges from fluid and solid mechanics. We conduct extensive experiments, assessing the operators' capabilities in learning, zero-shot super-resolution, data efficiency, robustness to noise, and computational efficiency. 
Interestingly, our findings highlight that current operators struggle with the newer mechanics datasets, motivating the need for more robust neural operators. All the datasets and \footnotemark[3]{}codes will be shared in an easy-to-use fashion for the scientific community. 
We hope this resource will be an impetus for accelerated progress and exploration in modeling dynamical systems. 
\end{abstract}

\section{Introduction}
Nature is in a continuous state of evolution. ``Rules'' governing the time evolution of systems in nature, also known as dynamics, can be captured mathematically through partial differential equations (PDEs). 
In the realm of science and engineering, PDEs are widely used to model and study several challenging real-world systems, such as fluid flow, deformation of solids, plasma dynamics, robotics, mechanics, and weather forecasting, to name a few \citep{debnath2005nonlinear, nakamura1977computational, robert2007partial}. Due to their highly non-linear and coupled nature, these PDEs can be solved analytically only for trivial or model systems. Thus, accurate numerical solutions for the PDEs are the cornerstone in advancing scientific discovery. Traditionally, the PDEs are solved using classical numerical methods such as finite difference, finite volume, or finite element methods \citep{sewell2012analysis}. However, these numerical methods exhibit major challenges in realistic systems in terms of system size, timescales, and numerical instabilities. Specifically, simulating the systems for longer timescale or for large domains is extremely computationally intensive to the extent that performing them in real-time for decision-making is a major challenge. Further, in the case of large/highly non-linear fields, these simulations often exhibit numerical instabilities, rendering them ineffective. \citep{solin2005partial}
The recent surge in artificial intelligence-based approaches suggests that neural models can efficiently capture continuous dynamical systems in a data-driven fashion \citep{brunton2022data}. These models are extremely time-efficient in comparison to traditional solvers and can capture highly non-linear input-output relationships. Earlier approaches in this direction relied directly on learning the input-output map through multilayer perceptrons (MLPs), convolutional neural networks, or graph neural networks. However, these approaches faced challenges in terms of generalizing to unseen initial or boundary conditions, geometries, or resolutions. This could be attributed to the fact that the neural models essentially learn the input-output relationship in a finite-dimensional approximation. To address this challenge, a seminal theory, extending the universal approximation theorem of neural networks~\citep{cybenko1989approximation} to neural operators was proposed, namely, the universal operator approximation theory~\citep{chen1995universal}. This theory unveiled the neural networks' prowess in handling infinite-dimensional inputs and outputs. 
\FloatBarrier
\graphicspath{ {././} }
\begin{figure}[!t] 
    \centering
    \subfloat{%
        \includegraphics[width=0.33\textwidth]{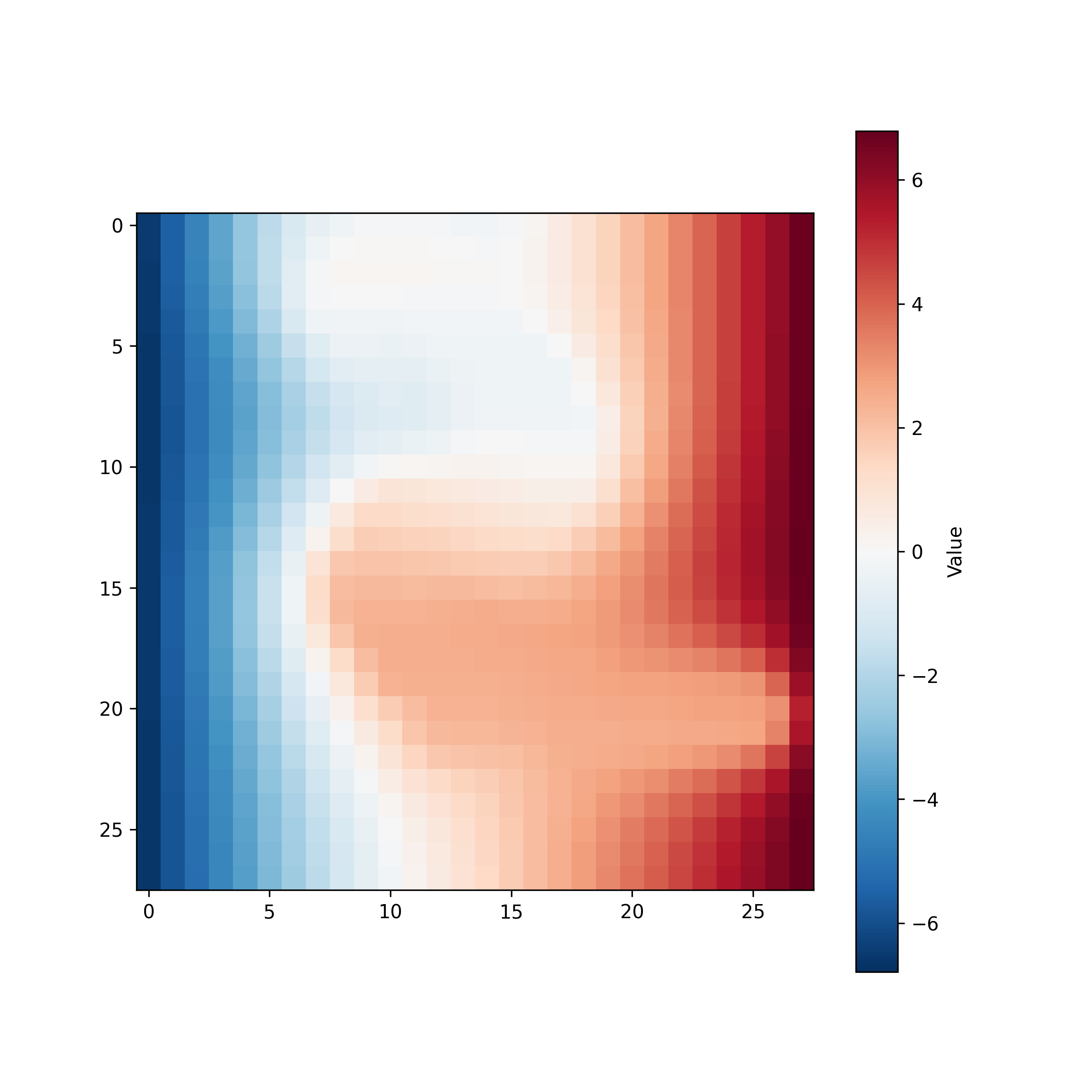}%
        }%
    \subfloat{%
        \includegraphics[width=0.33\textwidth]{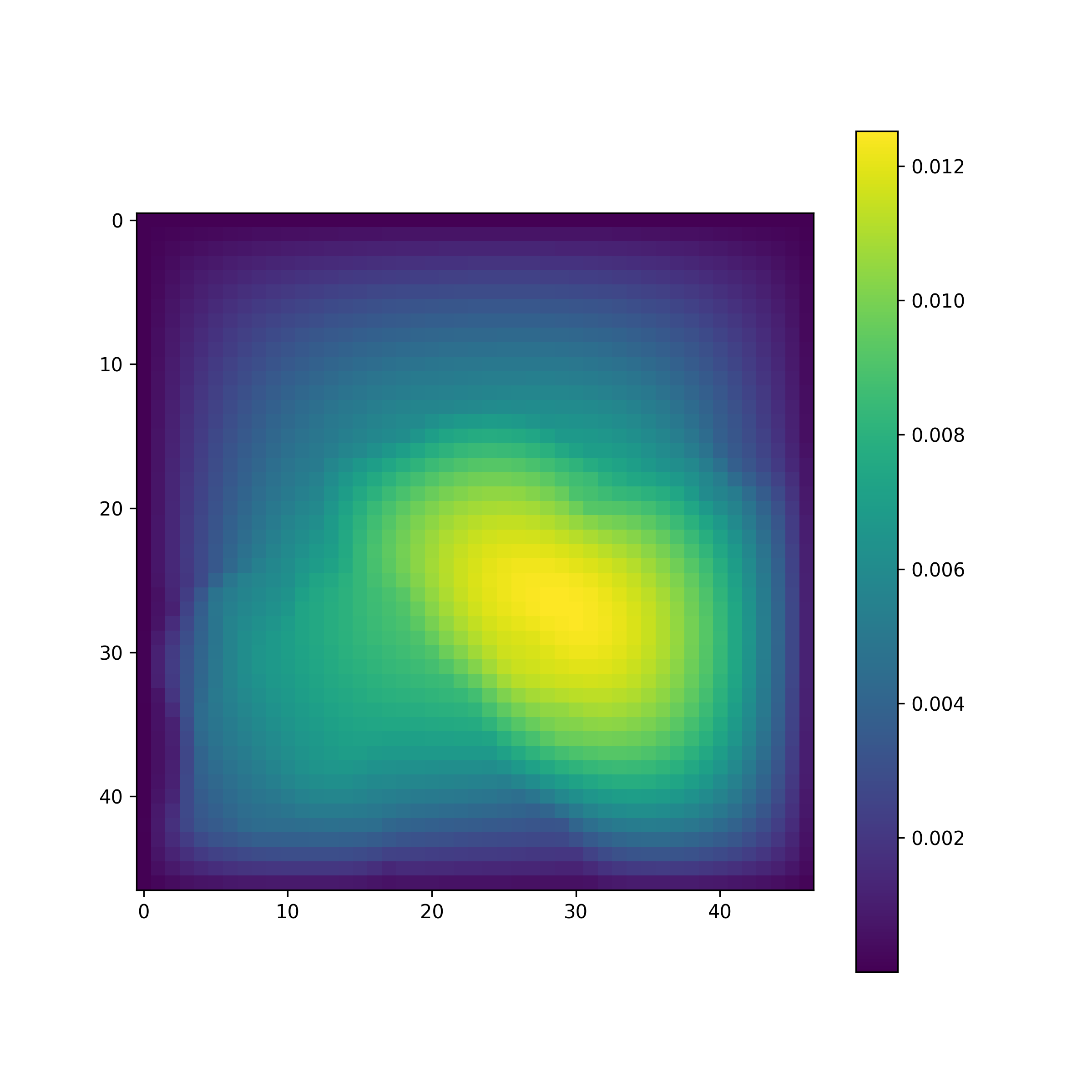}%
        }%
    \subfloat{%
        \includegraphics[width=0.33\textwidth]{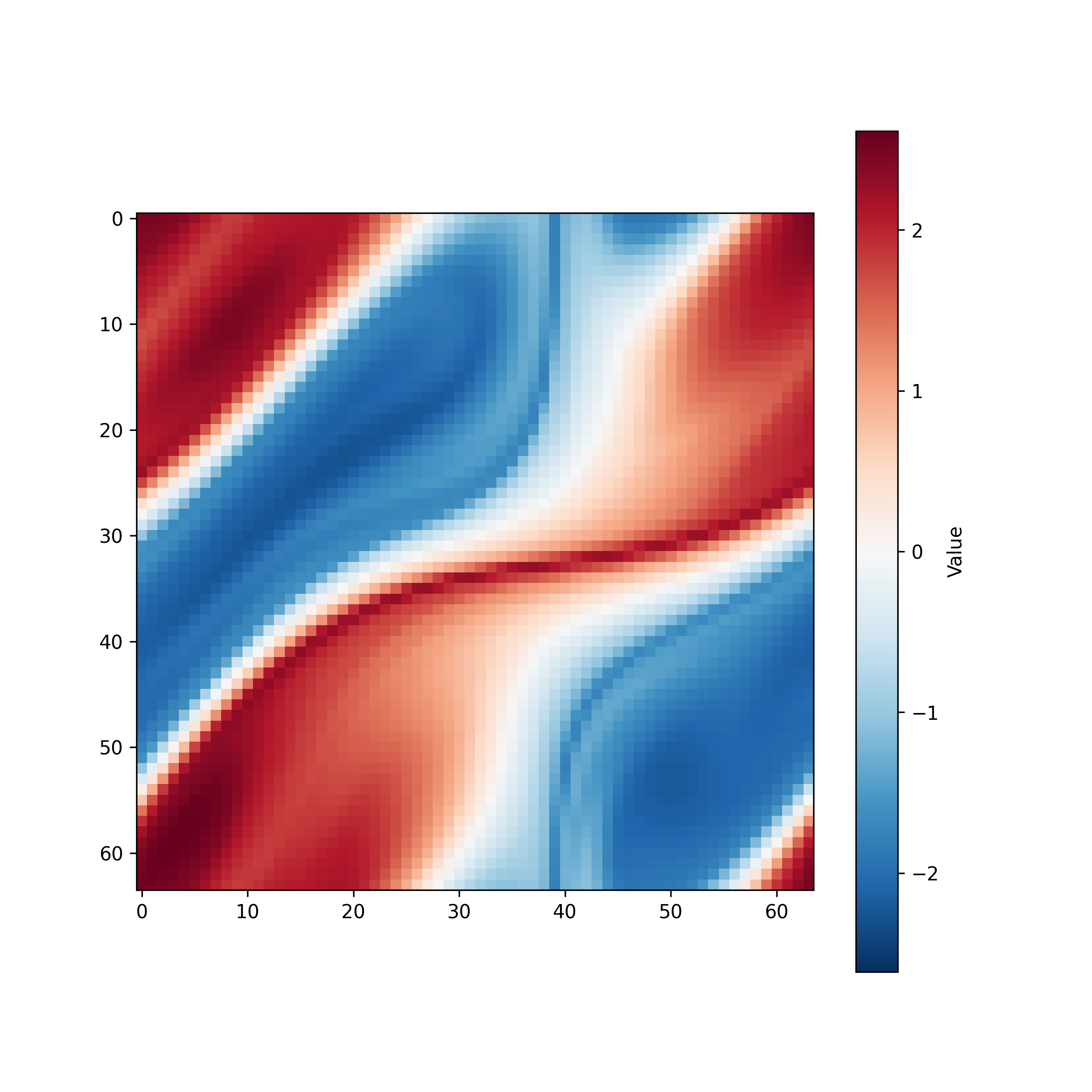}%
        }%
    \caption{A glimpse of the data diversity under consideration. From left to right, we present visual representations of three distinct datasets — Biaxial, Darcy Flow, and Navier-Stokes.}
    \label{fig:fig1}
    \vspace{-0.1in}
\end{figure}
\FloatBarrier

Theoretically, directly learning the solution operator through specialized neural network architectures offers several key advantages. (i) They can directly learn input-output function mappings from data,  thereby obviating the necessity for prior knowledge of the underlying PDE. (ii) They offer significantly improved time efficiency compared to traditional numerical solvers. (iii) They exhibit zero-shot generalizability, extending their applicability to systems of larger scale and complexity than those encompassed within the training dataset. (iv) They provide superior approximations of the solution operator compared to existing neural architectures, spanning from feed-forward networks to specialized models like convolutional networks and conditional generative adversarial networks (GANs). Thus, the neural operators attempt \citep{kovachki2021neural} to combine the best of both data-driven and physics-based numerical models.

This motivated the exploration of neural operator architectures \citep{bhattacharya2021model}, \citep{nelsen2021random},  capable of directly learning the solution operator. For instance, consider {\deeponet} \citep{lu2021learning}, which leverages the universal approximation theorem introduced by Chen and Chen to directly address PDEs. On a different front, {\fno} \citep{li2020fourier}, one of the most widely used Neural Operators, focuses on parameterizing the integral kernel within Fourier space. Moreover, a noteworthy study \citep{cao2021choose} highlights the notion that all transformers are essentially operators. This insight has sparked endeavors to create operator transformers. Given their proven effectiveness in sequence-to-sequence learning tasks, these transformer-based designs open avenues for enhancing the approximation of spatiotemporal PDEs. Prior studies, such as those by \citep{hao2022physics}, have delved into the realm of PINNs \citep{raissi2019physics} and some neural operator architectures, like {\deeponet}, {\fno}, and their variants. However, unlike fields like computer vision, comprehensive comparative evaluations of these neural operators are absent. Moreover, due to the variations and incompatibilities in architectures, a direct comparison of all these architectures is extremely cumbersome. Such evaluations are pivotal to discerning the distinctive advantages of diverse architectural paradigms, especially when addressing equations from a wide spectrum of scientific domains.

This study aims to bridge this gap by rigorously evaluating data-driven models that encompass a wide range of classes and methods, including the foundational deep operator regression model, frequency domain parameterization models, and transformer-based architectures, to achieve state-of-the-art performance comparison on selected PDE datasets. Moreover, we integrate conventional neural architectures to underscore the merits of PDE-specialized structures. Our dataset selection is methodical, designed to challenge each model with equations from various scientific disciplines. We incorporate four prevalent equations from fluid dynamics and four standard differential equations from solid mechanics into the neural operator domain, ensuring a holistic comparison within the realm of neural operators.

\textbf{Our Contribution:} In this work, we critically analyze 11 data-driven models, including operators and transformers, on 8 PDE datasets. The major contributions of our research are as follows:
\begin{enumerate}
    \item {\bf \cod:} We present a package that allows seamless analysis of several data-driven approaches on PDEs. We thoroughly assess state-of-the-art data-driven neural models for solving PDE datasets across diverse scientific realms, such as fluid and solid mechanics, shedding light on their precision and efficacy.
    \item {\bf Super-resolution:} We analyze the ability of neural operators' to generalize to systems of different resolutions than that of their training sets' discretizations.
    \item {\bf Data efficiency and robustness to noise:} We critically assess the efficiency of these models to learn from small amounts of data or noisy data. This is an important aspect since the data available can be scarce and noisy in practical applications.
    \item {\bf Out-of-distribution task:} A novel task to gain insights into what these models are truly learning to determine whether the underlying operator is genuinely being learned or if the training dataset is simply being fitted. Two closely related {\stress} and {\strain} datasets are interchanged during training and testing to dig deeper into whether the solvers are actually operators.
\end{enumerate}

\section{Preliminaries}
This section provides a concise mathematical framework to illustrate how traditional PDE solving can be transitioned and addressed using data-driven methodologies via neural networks.
\begin{enumerate}
    \item \textbf{Function Domains}: Consider a bounded open set, represented as $\mathcal{D} \subset \mathbb{R}^d$. Within this domain, we define $\mathcal{F} = \mathcal{F}(\mathcal{D};\mathbb{R}^{d_f})$ and $\mathcal{G} = \mathcal{G}(\mathcal{D};\mathbb{R}^{d_g})$ as separable Banach spaces. These spaces correspond to input and output functions, which represent elements in $\mathbb{R}^{d_f}$ and $\mathbb{R}^{d_g}$, respectively.
    \item \textbf{The Solution Operator}: In our exploration, we introduce $T^\dagger: \mathcal{F} \rightarrow \mathcal{G}$, a mapping that is typically nonlinear. This mapping emerges as the solution operator for PDEs, playing a pivotal role in scientific computations.
    \item \textbf{Data Generation}: For training purposes, models utilize PDE datasets constructed as $\mathcal{D} = \{(\mathcal{F}_k, \mathcal{G}_k)\}_{1\leq k\leq D}$, where $\mathcal{G}_k = T^\dagger(\mathcal{F}_k)$. Given the inherent challenges in directly representing functions as inputs to neural networks, the functions are discretized using mesh generation algorithms \citep{tristano1998advancing} over domain $\mathcal{D}$. We sample both input and output functions on a uniform grid, as it ensures compatibility with all selected solvers. For the input function $\mathcal{F}_k$, we discretize it on the mesh $\{x_i \in \Omega\}_{1\leq i\leq R}$, and the discretized $\mathcal{F}_k$ is $\{(x_i, f_{ik})\}_{1\leq i\leq R}$, where $f_{ik} = \mathcal{F}_k(x_i)$. Similarly, For the solution function $\mathcal{G}_k$, we discretize it on the mesh $\{y_i \in \Omega\}_{1\leq i\leq R}$, and the discretized $\mathcal{G}_k$ is $\{(y_i, g_{ik})\}_{1\leq i\leq R}$, where $g_{ik} = \mathcal{G}_k(y_i)$. It's worth noting that models such as {\poddeeponet} and {\sno} utilize only the function values for representation, excluding grid locations from the model input.
    \item \textbf{Objective}: The overarching goal for each model is to craft an approximation of $T^\dagger$. This is achieved by developing a parametric mapping, denoted as $T: \mathcal{F} \times \Theta \rightarrow \mathcal{G}$ or, in an equivalent form, $T_\theta: \mathcal{F} \rightarrow \mathcal{G}$, where $\theta \in \Theta$. This mapping operates within a bounded parameter space, $\Theta$. 
    \item \textbf{Metric}: Evaluating the efficacy of the parametric mapping involves comparing its outputs, $T_\theta(\mathcal{F}_k)$ = \{$\widetilde{g}_{ik}\}_{1\leq i\leq R}$, with the actual data, aiming to minimize the relative L2 loss, given by:
    \begin{equation}
    \min_{\theta\in \Theta} \frac{1}{D} \sum_{k=1}^{D} \frac{1}{R} \frac{\left\| {T}_\theta(\mathcal{F}_k) - \{\widetilde{g}_{ik}\}_{1\leq i\leq R} \right\|_2^2}{\left\| \{\widetilde{g}_{ik}\}_{1\leq i\leq R} \right\|_2^2},
    \end{equation}
\end{enumerate}
Here, \( R \) denotes the function discretization parameter. 

\section{Model Architectures}
\begin{figure}[!t]
    \centering
    \includegraphics[width=\textwidth, trim={3cm 23cm 5cm 1.5cm}, clip]{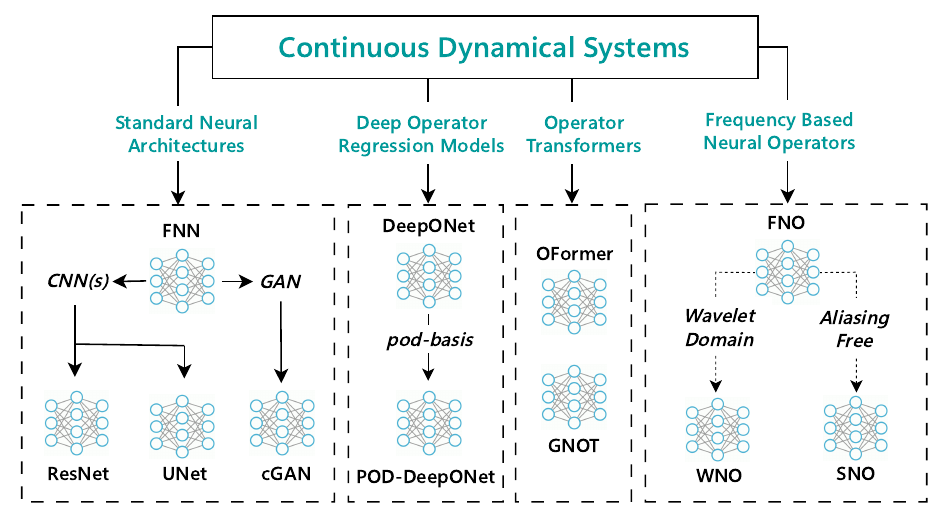}
    \caption{An overview of the various models being benchmarked and the relationship between them. The term `pod-basis' denotes the basis of the output function, derived directly from proper orthogonal decomposition, as opposed to being learned through a neural network.\label{fig:fig2}}
\end{figure}

\textbf{Standard Neural Network Architectures:}
This study encompasses a broad spectrum of architectures, as illustrated in Figure~\ref{fig:fig2}. The Feed-Forward Neural Network ({\fnn}) serves as the foundational component, distinguished by its pointwise configuration. Prevalent CNN-based architectures like {\unet}, {\resnet}, and {\cgan} are also incorporated. {\unet}, delineated in \citep{ronneberger2015u}, employs a U-shaped encoder-decoder design augmented by skip connections, facilitating the capture of both granular and abstract features. {\resnet}, described in \citep{jian2016deep}, consist of a series of residual blocks and are commonly used in computer vision tasks \citep{targ2016resnet}. Conditional Generative Adversarial Networks ({\cgan}), introduced in \citep{mirza2014conditional}, are an evolution of the GAN framework, facilitating conditional generation via the incorporation of label information in both the generator and discriminators.

\textbf{Deep Operator-Based Regression Models:}
Neural Operators represent a novel ML paradigm, predominantly employed in scientific machine learning to decipher PDEs. These operators discern mappings between infinite-dimensional Euclidean spaces, relying solely on data and remaining agnostic to the underlying PDE. Within this study, we delve into deep operator regression models. {\deeponet} bifurcates into two sub-networks: the branch net, which encodes the input function at fixed sensor locations, and the trunk net, encoding solution function locations \citep{lu2021learning}. The solution emerges from the inner product of the outputs from these nets. In {\poddeeponet}, the bases are determined by executing proper orthogonal decomposition (POD) on training data, replacing the self-learned basis of output functions \citep{lu2022comprehensive}. This POD basis forms the trunk net, leaving only the branch net as the trainable component, which discerns the coefficients of the POD basis.

\textbf{Frequency-Based Operators:}
Frequency-based solvers like {\fno} employ a finite-dimensional parameterization using truncated Fourier modes \citep{li2020fourier}. By integrating this with an integral operator restricted to convolution, and instantiated via a linear transformation in the Fourier domain, the {\fno} operator is conceived. {\wno}, or Wavelet Neural Operator, amalgamates the prowess of wavelets in time-frequency localization with an integral kernel. By learning the kernel in the wavelet domain, convolution operates on wavelet decomposition coefficients rather than direct physical space convolution \citep{tripura2022wavelet}. {\sno}, the Spectral Neural Operator, addresses the often-overlooked aliasing error in the Fourier Neural Operator. By representing both input and output functions using coefficients in truncated Fourier or Chebyshev series, SNO offers an aliasing-free approach \citep{fanaskov2022spectral}. Any transformation between these coefficients can be executed using neural networks, and methods employing these series are termed spectral neural operators. In their approach, a straightforward feed-forward neural network architecture in the complex domain is utilized.

\textbf{Transformer Operators:}
"{\gnot} introduces the Heterogeneous Normalized (linear) Attention (HNA) block and a geometric gating mechanism, specifically tailored for enhanced performance on PDE datasets \citep{hao2023gnot}. This architecture effectively performs a soft domain decomposition \citep{jagtap2021extended}, treating each decomposed domain independently and subsequently integrating them using a mixture-of-experts approach to predict the underlying truth. This design allows {\gnot} to serve as a versatile operator adept at handling a variety of PDE types. In contrast, the {\oformer} model builds upon the seminal work presented in \citep{vaswani2017attention}. It incorporates random Fourier projection to counteract spectral bias, thereby enhancing its efficacy on PDEs \cite{li2022transformer}. 

\section{Datasets}
Here, we briefly describe the 8 datasets used in the present. While previous approaches have mostly focussed on fluid datasets, here we present 4 datasets on fluid flow and 4 on the deformation of solids; for complete dataset details, refer \ref{detailed description}, \ref{mechanical}.
\begin{enumerate}
    \item \textbf{{\burgers}}:
    This dataset models the one-dimensional flow of a viscous fluid. The input is the fluid's initial velocity distribution at time $t = 0$, and the output is the fluid's velocity at a time $t > 0$~\cite{takamoto2022pdebench}.
    
    \item \textbf{{\darcy}}:
    The Darcy Flow dataset describes the steady-state flow of a fluid through a porous medium in two dimensions. The input is the spatial distribution of the medium's resistance to flow (viscosity), and the output is the fluid's velocity distribution across the domain at steady-state~\citep{takamoto2022pdebench}.
    
    \item \textbf{{\navierstokes}}:
    This dataset models the time evolution of a 2D viscous, incompressible fluid. The input includes the fluid's initial swirling motion (vorticity) and external forces acting on the fluid. The output is the fluid's velocity distribution over a specified time period~\citep{takamoto2022pdebench}.

    \item \textbf{{\shallowwater}}:
    The shallow-water equations simulate the behavior of water that flows over a shallow surface in 2D. The input consists of the initial water depth and velocity distribution, and the output predicts the water flow dynamics in response to gravitational forces and varying underwater terrain (bathymetry)~\citep{takamoto2022pdebench}.

    \item \textbf{{\stress}}:
    This dataset models the stress distribution in a 2D binary composite material subjected to mode-I tensile loading. The input is the material microstructure (distribution of two materials), and the output is the stress field ({\stress}) distribution of the digital composite \citep{mehran2022learning}.

    \item \textbf{{\strain}}:
    The strain dataset describes the deformation of a 2D binary composite material subjected to mode-I tensilie loading. The input is the material microstructure and the output is the resulting strain fields ({\strain})~\citep{mehran2022learning}.

    \item \textbf{{\shear}}: 
    Part of the mechanical MNIST collection, this dataset simulates the deformation of a heterogeneous material block when forces are applied parallel to its surface ({\shear}). The input is the material microstructure, and the output captures element-wise displacements subjected to shear loading \citep{lejeune2020mechanical}.

    \item \textbf{{\biaxial}}:
    Another subset of the mechanical MNIST experiments, this dataset models the material's response when stretched equally in two perpendicular directions (equibiaxial loading). The input is the material microsturcture, and the output records the full field displacement under {\biaxial} stretching \citep{lejeune2020mechanical}.
\end{enumerate}

\section{Benchmarking Results}
\begin{table}
\centering
\resizebox{\columnwidth}{!}{%
\begin{tabular}{@{}l|cccccccc@{}}
\toprule
{\color[HTML]{000000} } & \multicolumn{8}{c}{{\color[HTML]{000000} Datasets}}                                                                                                  \\ \midrule
Models                  & \textbf{\burgers}                                    & \textbf{\darcy}                                    & \textbf{\navierstokes}                             & \textbf{\shallowwater}                              & \textbf{\stress}                                   & \textbf{\strain}                                   & \textbf{\shear}                                    & \textbf{\biaxial}                                  \\ \midrule
\fnn                     & $5.853_{\pm1.416}$                                 & $3.47_{\pm0.14}$                                 & $34.77_{\pm0.19}$                                 & $2.424_{\pm0.656}$                                 & $25.69_{\pm0.59}$                                & $23.09_{\pm7.08}$                                & {\color[HTML]{3531FF} $1.11_{\pm0.06}$}          & {\color[HTML]{3531FF} $3.69_{\pm0.01}$}          \\
\resnet                  & $11.327_{\pm1.208}$                                & $5.14_{\pm0.23}$                                 & $29.52_{\pm0.14}$                                 & {\color[HTML]{3166FF}$0.287_{\pm0.093}$}                               & $20.05_{\pm0.19}$                                & $14.64_{\pm0.31}$                                & $3.02_{\pm0.95}$                                 & $13.58_{\pm2.67}$                                \\
\unet                    & $30.870_{\pm2.000}$                                & $2.10_{\pm0.08}$                                 & $24.02_{\pm0.95}$                                 & $0.295_{\pm0.097}$                                 & {\color[HTML]{3166FF} $10.57_{\pm0.19}$}         & {\color[HTML]{3166FF} $9.05_{\pm0.33}$}          & $7.09_{\pm0.46}$                                 & $16.63_{\pm2.30}$                                \\
\cgan                    & $34.906_{\pm0.506}$                                & {\color[HTML]{3531FF} $1.88_{\pm0.04}$}          & $24.00_{\pm0.48}$                                 & $0.291_{\pm0.027}$                                 & {\color[HTML]{3531FF} \boldsymbol{$6.66_{\pm0.84}$}} & {\color[HTML]{3531FF} $6.12_{\pm0.80}$}          & $5.63_{\pm0.50}$                                 & $15.74_{\pm1.40}$                                \\
\fno                     & {\color[HTML]{3531FF} \boldsymbol{$0.160_{\pm0.004}$}} & {\color[HTML]{3531FF} \boldsymbol{$1.08_{\pm0.06}$}} & {\color[HTML]{3531FF} $14.13_{\pm0.34}$} & {\color[HTML]{3531FF} $0.128_{\pm0.018}$}          & {\color[HTML]{3531FF} $8.08_{\pm0.15}$}          & {\color[HTML]{3531FF} \boldsymbol{$5.61_{\pm0.23}$}} & {\color[HTML]{3166FF} $2.25_{\pm1.14}$}          & {\color[HTML]{3166FF} $7.40_{\pm1.91}$}          \\
\wno                     & $7.332_{\pm0.307}$                                 & $2.23_{\pm0.14}$                                 & $37.08_{\pm1.23}$                                 & $0.572_{\pm0.036}$                                 & $17.24_{\pm0.46}$                                & $12.05_{\pm0.26}$                                & $4.37_{\pm0.08}$                                 & $22.22_{\pm2.86}$                                \\
\sno                     & $40.623_{\pm8.437}$                                & $8.55_{\pm1.03}$                                 & $98.46_{\pm0.25}$                                 & $94.891_{\pm0.060}$                                & $51.31_{\pm0.01}$                                & $62.34_{\pm1.17}$                                & $4.37_{\pm0.87}$                                 & $21.93_{\pm0.57}$                                \\
\deeponet                & $10.561_{\pm1.182}$                                & $4.27_{\pm0.24}$                                 & $55.48_{\pm1.06}$                                 & $8.602_{\pm0.431}$                                 & $24.59_{\pm0.98}$                                & $23.75_{\pm0.20}$                                & $2.85_{\pm0.18}$                                 & $8.28_{\pm0.37}$                                 \\
\poddeeponet            & $3.999_{\pm0.654}$                                 & $3.43_{\pm0.04}$                                 & $33.37_{\pm1.30}$                                 & $1.503_{\pm0.145}$                                 & $29.63_{\pm0.52}$                                & $18.31_{\pm1.17}$                                & $4.14_{\pm0.44}$                                 & $30.46_{\pm0.59}$                                \\
\oformer                 & {\color[HTML]{3531FF} $0.165_{\pm0.016}$}          & $3.21_{\pm0.06}$                                 & {\color[HTML]{3531FF} \boldsymbol{$10.97_{\pm3.03}$}}          &        $6.597_{\pm0.352}$                                             & $27.33_{\pm0.28}$                                & $25.08_{\pm1.36}$                                & $41.75_{\pm0.19}$                                & $61.16_{\pm0.49}$                                \\
\gnot                    & {\color[HTML]{3166FF} $0.677_{\pm0.021}$}          & {\color[HTML]{3166FF} $2.04_{\pm0.05}$}          & {\color[HTML]{3166FF} $23.73_{\pm0.97}$}          & {\color[HTML]{3531FF} \boldsymbol{$0.102_{\pm0.007}$}} & $13.02_{\pm0.81}$                                & $9.99_{\pm0.62}$                                 & {\color[HTML]{3531FF} \boldsymbol{$0.43_{\pm0.02}$}} & {\color[HTML]{3531FF} \boldsymbol{$0.71_{\pm0.04}$}} \\ \bottomrule
\end{tabular}%
}
\vspace{0.5cm}
\caption{Performance of different models across diverse datasets from distinct domains. The Relative L2 Error, expressed as ($\times 10^{-2}$), is presented as the evaluation metric. Lower scores denote better performance. The optimal outcomes are highlighted in bold and dark blue, followed by the second-best in dark blue, and the third-best in light blue.\label{tab:table1}}
\end{table}

We present the results of rigorous experimentation on PDE solvers across six tasks, each designed to showcase the unique capabilities and strengths of the models. The diversity of the selected PDEs, sourced from \citep{takamoto2022pdebench}, \citep{lejeune2020mechanical}, and \citep{mehran2022learning}, encompasses both time-dependent and time-independent challenges, capturing the intrinsic computational complexity inherent to these tasks. Specifically, the experiments conducted on novel mechanical datasets not previously encountered by the solvers, offer invaluable insights for the broader scientific community.

In alignment with established experimental protocols, the dataset was split as follows: $\sim 80\%$ for training, $\sim 10\%$ for validation, and $\sim 10\%$ for testing. Our ensemble training methodology ensured a level playing field for each operator by defining a hyperparameter range and selecting the best subset for experimentation. Model optimization was achieved using the Adam \citep{kingma2014adam} and AdamW \citep{loshchilov2017decoupled} optimizers. Depending on the specific task, we employed either step-wise or cycle learning rate scheduling \citep{smith2019super}, with the optimal learning rate chosen from the set $\sim$ \(\{10^{-1}, 10^{-2}, 10^{-3}, 10^{-4}, 10^{-5}\}\).

The training was conducted under optimal hyperparameter configurations, introducing variability through distinct random seeds and data splits. All experiments adhered to a fixed batch size of 20 and were executed on 1 $\sim$ 8 NVIDIA A6000 GPUs, with memory capacities of 48 GBs. To ensure fairness and accuracy in results, each experiment was replicated thrice with different seeds. We report the mean and deviation in Relative L2 Error. 

\subsection {Accuracy}
Table ~\ref{tab:table1} shows the performance of the models on the 8 datasets. {\fno} architecture stands out on all datasets, consistently delivering results among the best three. Its strength lies in its transformation in the frequency space. By capturing and transforming the lower frequencies present in the data, the {\fno} can approximate the solution operators of scientific PDEs. This approach, which uses the integral kernel in the Fourier space, facilitates a robust mapping between input and output function spaces, making it particularly adept at handling the complexities of the datasets in this study. Following \fno, {\gnot}, employing a mixture-of-experts approach, showcases exemplary performance on most (6/8) datasets. Its unique soft domain decomposition technique divides the problem into multiple scales, allowing it to capture diverse features of the underlying PDE. Each expert or head in the model focuses on a different aspect of the PDE, and their combined insights lead to a comprehensive understanding, especially for challenging datasets like {\shear} and {\biaxial}.

The {\oformer{}} architecture, that employs an innovative approach to solving spatio-temporal PDEs, exhibits best results in \navierstokes{} dataset. By unrolling in the time dimension and initiating with a reduced rollout ratio, it efficiently forwards the time step dynamics in the latent space. This method conserves significant space during training on time-dependent datasets. It also significantly enhances the approximation while tested on one of the most complex dataset in the study, the time-dependent navier stokes PDE. 

Interestingly, most models, with the notable exception of {\gnot}, struggle to accurately learn the underlying PDE for the {\biaxial} and {\shear} datasets. Here, the simpler {\fnn} architecture demonstrates significant proficiency in learning these datasets. Interestingly, architectures like {\cgan}, originally designed for 2D image data analysis with its U-Net encoder, demonstrate impressive performance across tasks. This underscores the versatility of such architectures, even when they aren't explicitly designed as operators.

\subsection {Robustness to Noise}

In practical applications, it's common to encounter noise in measurements. To understand how various neural operators handle such real-world challenges, we simulated conditions with noisy data. During our testing phase, we intentionally introduced corrupted input function data to each model. The goal was to see how well these models could predict the ground truth amidst this noise.

\begin{figure}[!t] 
    \subfloat{%
        \includegraphics[width=0.4\textwidth,
        trim={0.25cm 0.6cm 5.5cm 0.2cm}, clip]{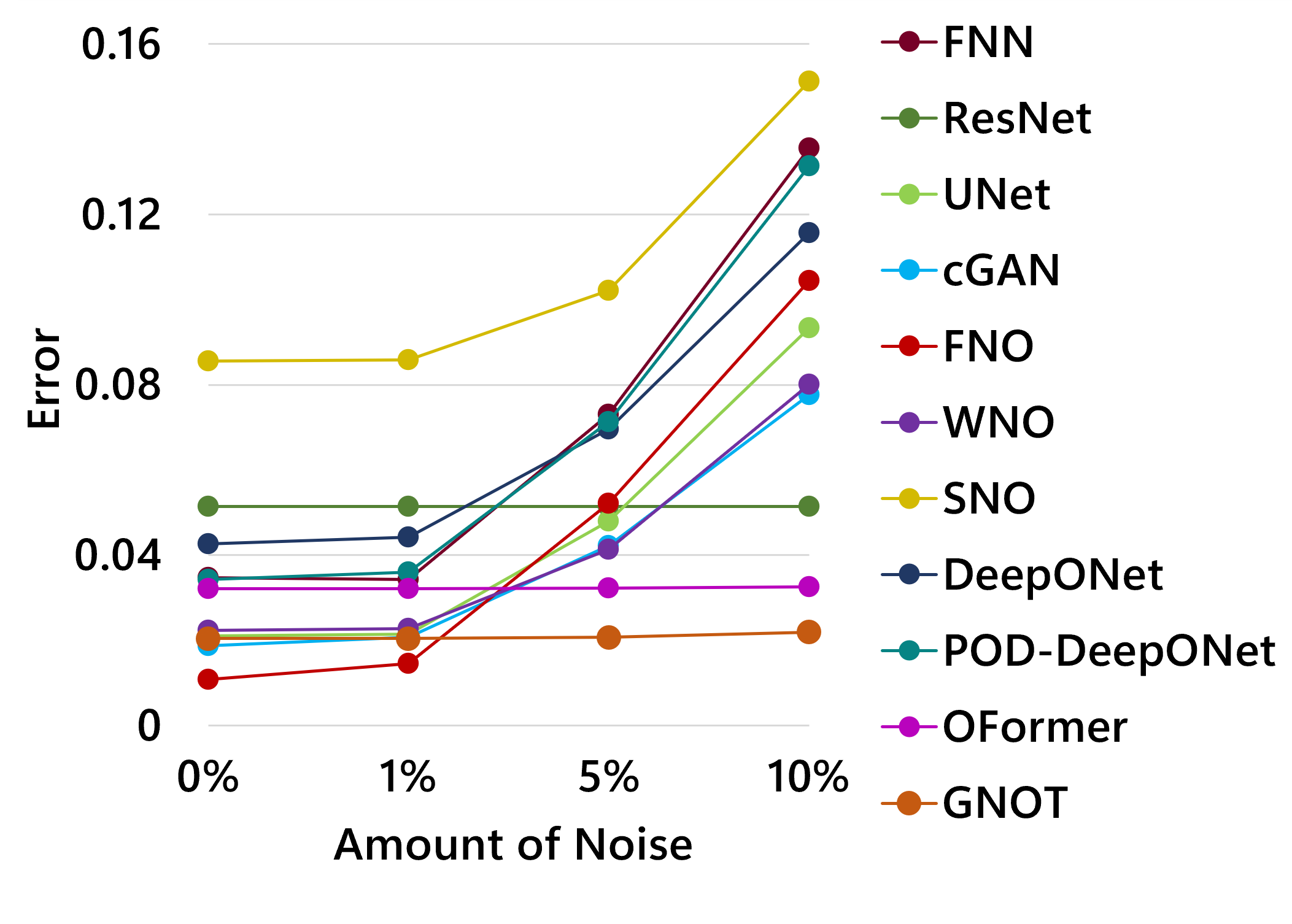}%
        }%
    \subfloat{%
        \includegraphics[width=0.4\textwidth, trim={0.25cm 0.618cm 5.5cm 0.2cm}, clip]{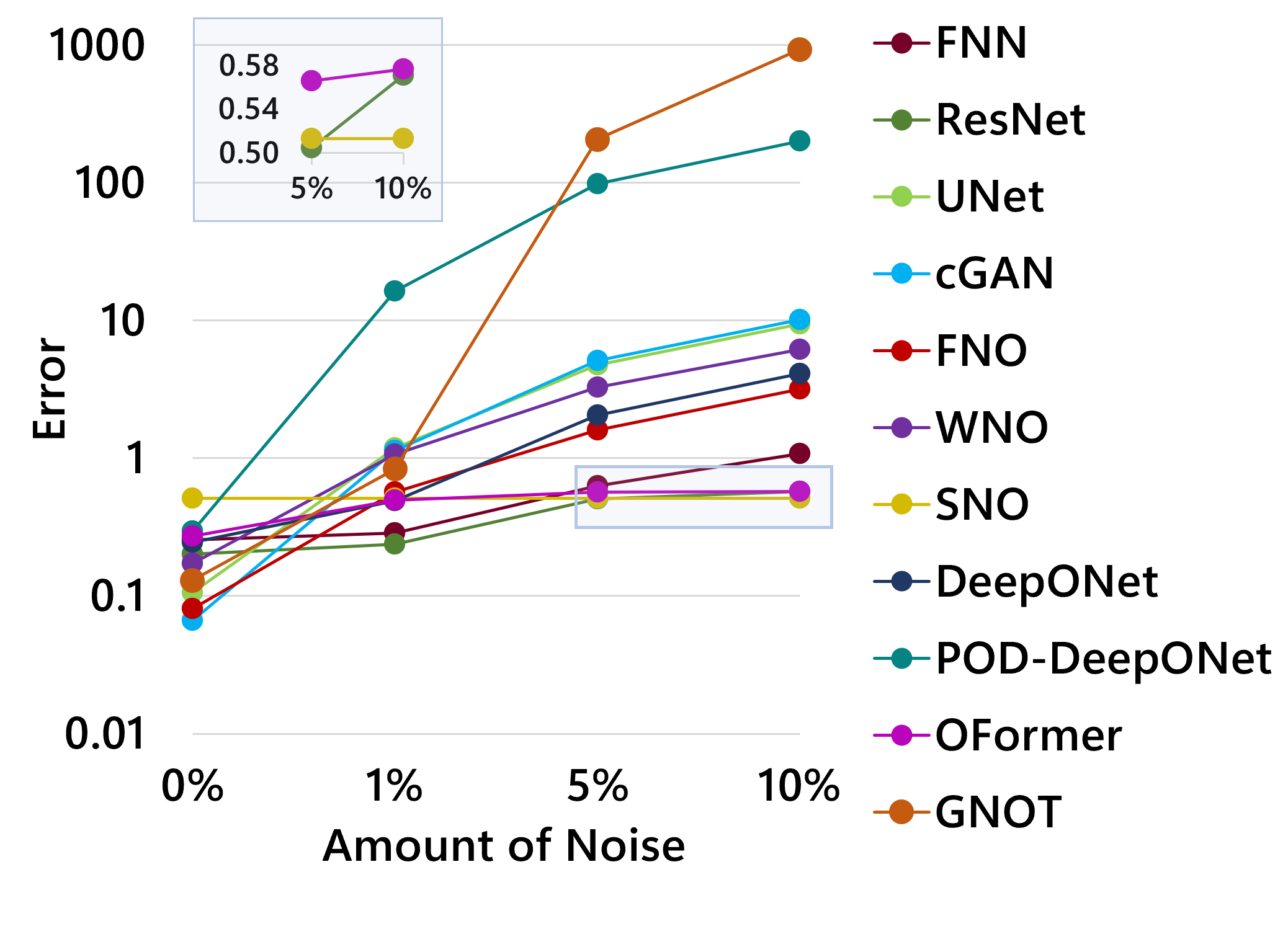}%
        }%
    \subfloat{%
        \includegraphics[width=0.2\textwidth, trim={10.5cm 0.6cm 0.2cm 0.2cm}, clip]{./noise2zoomwin.png}%
        }%
    \caption{Robustness Analysis Against Noise: Performance metrics, in terms of Relative L2 Error, are presented for models subjected to random Gaussian noise. The evaluation encompasses the {\darcy} dataset (left) and the {\stress} dataset (right). The right diagram provides a detailed comparison of the most noise-resilient models on the {\stress} dataset, specifically {\sno}, {\oformer}, and {\resnet}.\label{fig:fig3}}
\end{figure}

Figure~\ref{fig:fig3} shows the performance of the models on noisy data. Transformer-based architectures have shown commendable performance on the {\darcy} dataset. Even when noise is introduced, these models continue to perform well. However, their resilience is tested when faced with the {\stress} dataset. In scenarios where they already find it challenging to learn the underlying PDEs, the addition of noise exacerbates their performance issues, causing a noticeable decline in accuracy.

On the other hand, the spectral neural operator {\sno} shows superior robustness to noisy data. While its performance in a noise-free environment is far from best, it performs remarkably when exposed to noisy dataset, especially on the {\stress} dataset. This resilience can be attributed to its unique approach: unlike other frequency-based methods that transition between the time and frequency domains, {\sno} exclusively processes data in the frequency domain. This design choice allows it to effectively filter out noise, identifying it as a high-frequency disturbance, before it even begins its prediction process.
\FloatBarrier
\graphicspath{ {././} }
\FloatBarrier

\subsection {Data Efficiency}

For the data-efficiency experiments, we utilized the {\darcy} dataset with 1700 samples of $47\times47$ dimensions. To assess the data efficiency of the models, we trained all models on reduced subsets: 25\% (425 samples) and 50\% (850 samples) of the original dataset, while maintaining the same testing and validation datasets.

The exceptional performance of frequency-based methods, notably {\fno} and {\wno}, even with limited data, is rooted in their operation within the frequency domain (see Table ~\ref{tab:table2}). The notable capability of these methods to capture the essential dynamics of the underlying PDEs through the lower frequencies present in the data enables data-efficient learning, a crucial feature for realistic data where the number of  observations may be limited.

\begin{table}
\centering
\resizebox{\columnwidth}{!}{%
\begin{tabular}{@{}l|lllllllllll@{}}
\toprule
 & \multicolumn{11}{c}{Models}                                                                                                                                                                                                                                                    \\ \midrule
Dataset Size & \fnn                                   & \resnet            & \unet              & \cgan              & \fno               & \wno               & \sno               & \deeponet          & \begin{tabular}[c]{@{}l@{}}\textsc{POD-}\\ \deeponet\end{tabular} & \oformer           & \gnot              \\ \midrule
25\%         &       $4.80_{\pm0.27}$                                &      $6.23_{\pm0.23}$             &   {\color[HTML]{3531FF}$2.60_{\pm0.14}$ }               &      $3.28_{\pm0.13}$             &  {\color[HTML]{3531FF} \boldsymbol{$1.87_{\pm0.13}$}}                 &     {\color[HTML]{3166FF}$2.94_{\pm0.20}$}             &     $24.70_{\pm1.08}$              &     $7.50_{\pm0.45}$              &  $5.09_{\pm0.20}$                                                       &     $3.94_{\pm0.13}$              &       $3.61_{\pm0.20}$            \\
50\%         &          $3.95_{\pm0.24}$                             &       $5.20_{\pm0.29}$            &    {\color[HTML]{3531FF}$2.10_{\pm0.11}$ }              &   $2.54_{\pm0.13}$                &  {\color[HTML]{3531FF} \boldsymbol{$1.32_{\pm0.10}$}}                 &   {\color[HTML]{3166FF}$2.37_{\pm0.18}$}                &   $24.70_{\pm1.12}$                &    $6.15_{\pm0.41}$               &                               $4.17_{\pm0.28}$                          &      $3.32_{\pm0.08}$             & $2.70_{\pm0.13}$                  \\
100\%        & \multicolumn{1}{c}{$3.47_{\pm0.14}$} & $5.14_{\pm0.23}$ & $2.10_{\pm0.08}$ & {\color[HTML]{3531FF} $1.88_{\pm0.04}$} & {\color[HTML]{3531FF} \boldsymbol{$1.08_{\pm0.06}$}} & $2.23_{\pm0.14}$ & $8.55_{\pm1.03}$ & $4.27_{\pm0.24}$ & $3.43_{\pm0.04}$                                       & $3.21_{\pm0.06}$ & {\color[HTML]{3166FF}$2.04_{\pm0.05}$} \\ \bottomrule
\end{tabular}%
}
\vspace{0.1in}
\caption{Data-Efficiency Analysis: The Relative L2 Error (${\times} 10^{-2})$ is reported when trained with reduced subsets of 25\% and 50\% of the training dataset (left column). The testing and validation datasets remain consistent across all experiments.\label{tab:table2}}
\end{table}

\begin{table}
\centering
\resizebox{\columnwidth}{!}{%
\begin{tabular}{@{}ll|lclllllllll@{}}
\toprule
\multicolumn{2}{c|}{Dataset}               & \multicolumn{11}{c}{Models}                                                                                                              \\ \midrule
\multicolumn{1}{l|}{Train} &   \multicolumn{1}{l|}{Test}      & \fnn                                       & \resnet                                                        & \unet                                      & \cgan                                              & \fno                                               & \wno                & \sno                                                & \deeponet           & \begin{tabular}[c]{@{}l@{}}\textsc{POD-}\\ \deeponet\end{tabular} & \oformer                                   & \gnot               \\ \midrule
\multicolumn{1}{l|}{\stress} &
\multicolumn{1}{l|}{\stress} & $25.69_{\pm0.59}$                        & $20.05_{\pm0.19}$                                            & {\color[HTML]{3166FF} $10.57_{\pm0.19}$} & {\color[HTML]{3531FF} \boldsymbol{$6.66_{\pm0.84}$}} & {\color[HTML]{3531FF} $8.08_{\pm0.15}$}          & $17.24_{\pm0.46}$ & $51.31_{\pm0.01}$                                 & $24.59_{\pm0.98}$ & $29.63_{\pm0.52}$                                      & $27.33_{\pm0.28}$                        & $13.02_{\pm0.81}$ \\
 \multicolumn{1}{l|}{} & \multicolumn{1}{l|}{\strain}  & $91.11_{\pm0.04}$                        & {\color[HTML]{3166FF} $89.83_{\pm0.79}$}                     & $95.79_{\pm3.40}$                        & $95.34_{\pm1.57}$                                & $95.39_{\pm0.89}$                                & $93.71_{\pm4.97}$ & {\color[HTML]{3531FF} \boldsymbol{$62.36_{\pm0.46}$}} & $92.70_{\pm3.30}$ & $596.33_{\pm23.70}$                                    & {\color[HTML]{3531FF} $68.70_{\pm0.76}$} & $94.35_{\pm1.09}$ \\ \midrule
\multicolumn{1}{l|}{\strain} &
\multicolumn{1}{l|}{\strain} & $23.09_{\pm7.08}$                        & \multicolumn{1}{l}{$14.64_{\pm0.31}$}                        & {\color[HTML]{3166FF} $9.05_{\pm0.33}$}  & {\color[HTML]{3531FF} $6.12_{\pm0.80}$}          & {\color[HTML]{3531FF} \boldsymbol{$5.61_{\pm0.23}$}} & $12.05_{\pm0.26}$ & $62.34_{\pm1.17}$                                 & $23.75_{\pm0.20}$ & $18.31_{\pm1.17}$                                      & $25.08_{\pm1.36}$                        & $9.99_{\pm0.62}$  \\
\multicolumn{1}{l|}{} & \multicolumn{1}{l|}{\stress}  & {\color[HTML]{3531FF} $75.63_{\pm1.49}$} & \multicolumn{1}{l}{{\color[HTML]{3166FF} $77.29_{\pm0.72}$}} & $77.41_{\pm0.93}$                        & $79.49_{\pm1.07}$                                & $79.50_{\pm0.86}$                                & $80.56_{\pm1.27}$ & {\color[HTML]{3531FF} \boldsymbol{$51.65_{\pm1.10}$}} & $77.49_{\pm1.50}$ & $86.32_{\pm2.24}$                                      & $80.26_{\pm0.81}$                        & $80.24_{\pm0.95}$ \\ \bottomrule
\end{tabular}%
}
\vspace{0.15in}
\caption{Out-of-Distribution Evaluation: Models are trained on the {\stress} dataset and subsequently tested on both the {\stress} dataset and the out-of-distribution {\strain} dataset. The experiment is reciprocated with {\strain} as the training set. Relative L2 Error (${\times} 10^{-2}$) is reported, with top-performing results highlighted.\label{tab:table3}}
\end{table}

Transformer-based neural operator architectures have demonstrated potential in approximating operators. However, their efficacy diminishes when data is sparse. {\gnot}, which typically excels with a rich dataset, struggles to outperform even basic neural network architectures in a data-limited scenario. This trend underscores the inherent data dependency of transformer architectures, highlighting the challenges faced by many models, except frequency-based operators, when trained on limited data.

\subsection {Zero-shot Super-resolution}
Directly approximating the solution operator offers a theoretical advantage: the potential for a mesh invariant continuous dynamical system. Once trained, such a system can ideally maintain accuracy even when applied to much larger systems than those it was trained on. This capability is termed ``zero-shot super-resolution."
\columnratio{0.63}
\begin{paracol}{2}
\switchcolumn
        \begin{table}[!h]
        \centering
        \resizebox{0.3\textwidth}{!}{%
        \begin{tabular}{@{}l|ll@{}}
        \toprule
        \darcy & \multicolumn{2}{c}{Models}            \\ \midrule
        Resolution & \fno               & \gnot              \\ \midrule
        \boldsymbol{$47\times47$}      & \boldsymbol{$1.08_{\pm0.06}$} & \boldsymbol{$2.04_{\pm0.05}$} \\
        $64\times64$      &     $60.50_{\pm5.49}$              &   $55.32_{\pm5.65}$                \\
        $128\times128$    &     $59.99_{\pm5.48}$              &      $55.42_{\pm5.68}$             \\ \bottomrule
        \end{tabular}%
        }
        \vspace{0.1in}
        \caption{Zero-shot super-resolution. Comparing on various resolutions (left column) with corresponding model performance (right column). The original training resolution and its associated performance are highlighted in bold.}
        \label{tab:table4}
        \end{table}
        \vspace{-0.25in}
\switchcolumn

Note that \fno{} and \gnot{} enable zero shot super resolution without any modifications. Other models such as {\sno} and {\deeponet}, upon closer examination, reveals that they cannot have a straightforward application on zero-shot super-resolution. Instead, it leans on certain adjustments and workarounds to achieve the desired results. While these modifications might enable super-resolution in practice, they diverge from the concept of zero-shot super-resolution from an architectural perspective. Accordingly, for our evaluation, we consider only \fno{} and \gnot.  We trained both {\fno} and {\gnot} on the {\darcy} dataset at a resolution of $47\times47$. We then tested their performance on higher resolutions: $64\times64$ and $128\times128$.
\end{paracol}
As seen in Table ~\ref{tab:table4}, both models exhibited a rapid decline in performance as the resolution increased. While {\gnot} fared slightly better, its results were still not up to the mark.

\subsection {Out-of-distribution Generalization}
The equations for {\stress} and {\strain} are intrinsically linked, differing primarily by the coefficient of elasticity, commonly known as Young's modulus. Given that our training and testing processes utilize normalized data, it's reasonable to anticipate that the models trained on the {\stress} dataset should be adept at predicting strain in the material microstructures, and vice versa. This expectation is particularly true for neural operators, which are designed to grasp the underlying partial differential equations (PDEs) governing such relationships. Table ~\ref{tab:table3} shows the OOD evaluation on all the models. Interestingly, for \sno{}, the error on the strain test dataset remains consistent, whether it was trained on the strain dataset or the stress dataset. The same holds true when tested on the stress dataset. This consistency underscores \sno's ability to learn the underlying PDE. In stark contrast, other models don't exhibit this adaptability. Their accuracy levels decline when the testing set is swapped, indicating a potential limitation in their ability to generalize across closely related tasks.

\subsection {Time Efficiency}

\graphicspath{ {././} }
\FloatBarrier
\begin{figure}[!t]
    \subfloat{%
        \includegraphics[width=0.48\textwidth,trim={0 0.22cm 0 0},clip]{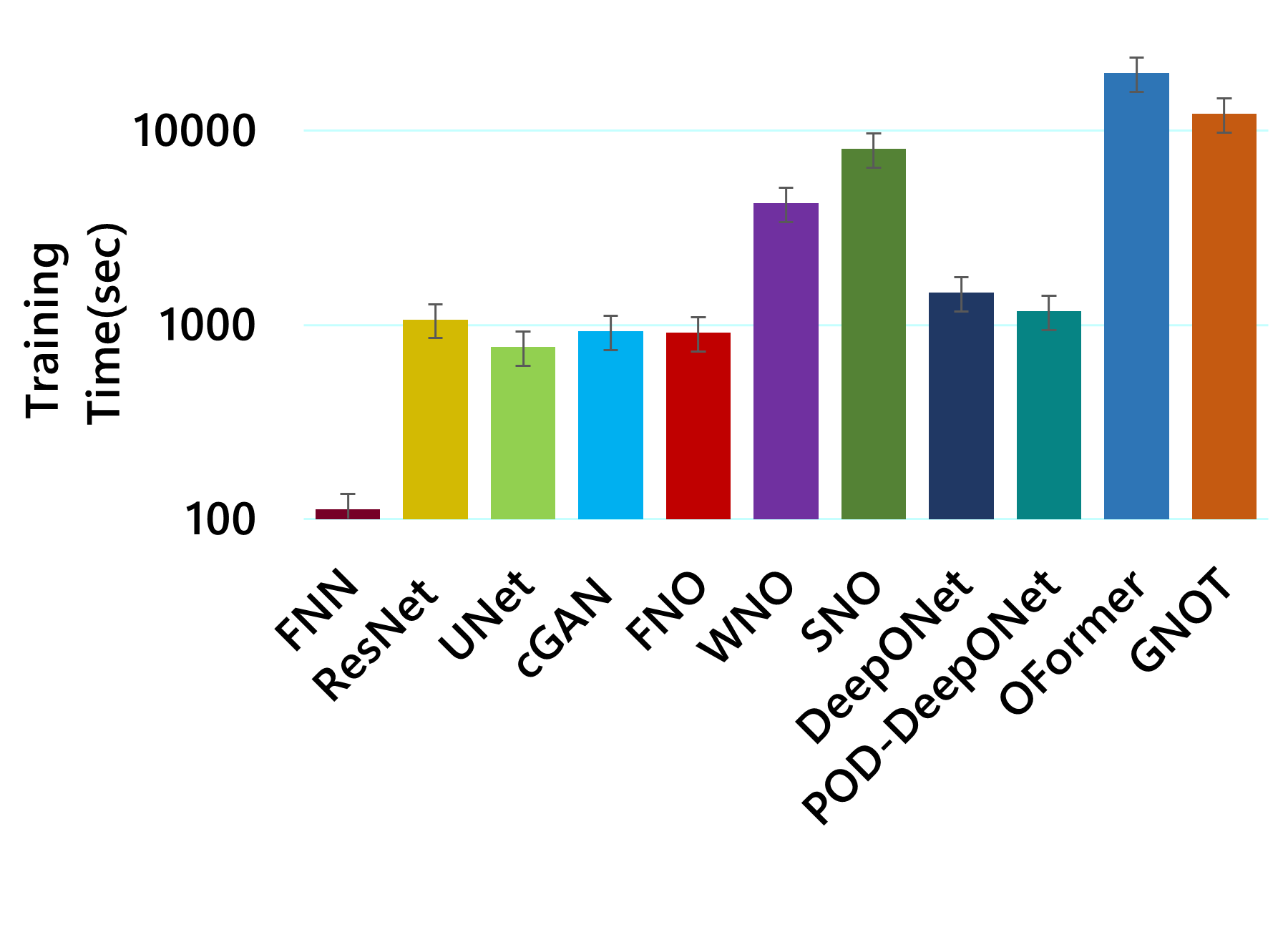}%
        }%
        \hspace{.02\textwidth}
    \subfloat{%
        \includegraphics[width=0.48\textwidth]{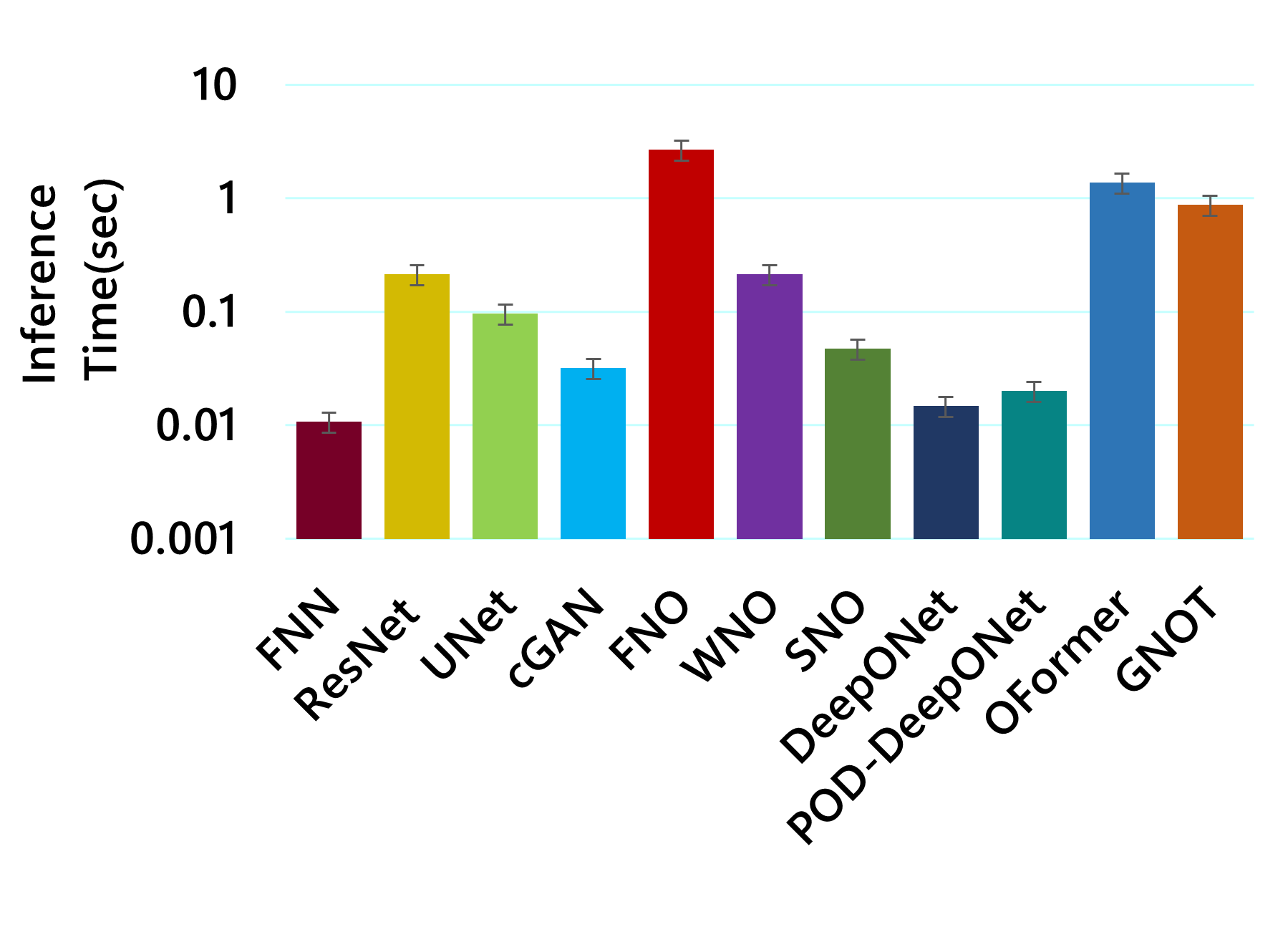}%
        }%
    \caption{Time Efficiency: We report the time taken by each model during training (on left) and inference time on test set (on right). Results are collected during the training on {\darcy} dataset.}
    \label{fig: fig4}
\end{figure}
\FloatBarrier
Neural Operators are gaining traction over traditional numerical solvers due to their promise of rapid inference once trained. For this assessment, we've bench-marked various continuous dynamical systems on two criteria: the duration required to train on the {\darcy} dataset and the time needed to predict output function values for 200 test samples, each mapped on a uniform $47\times47$ grid. As anticipated, the {\fnn}, with its straightforward architecture, stands out by requiring the least amount of time for both training and inference. However, when we delve into the other models, those based on deep operator regression methods, show training duration on par, with some of the complex but standard neural network architectures. For better visualization, see Figure \ref{fig: fig4}.
The narrative shifts when we consider inference time, a pivotal metric in practical applications. While {\fno} is relatively efficient during training, it, along with the transformer-based models, takes a longer inference stride. Though all these models show promising performance on different metrics, inference time efficiency remains a challenge for them. In stark contrast, most other models edge closer to offering real-time inference, highlighting the inherent time complexity trade-offs one must consider when opting for a particular neural operator.

\section{Concluding Insights}
The key insights drawn from this work are as follows.
\begin{enumerate}
    \item \textbf{Operator and transformer}: Both {\fno} and {\gnot} emerge as the superior models across various metrics, suggesting that the architectural novelty in these models can indeed capture the maps between infinite-dimensional functional spaces. 
    \item \textbf{Spectral resilience to noise and OOD}: Despite its underwhelming performance, {\sno} exhibits extreme resilience against noise. Similarly, \sno{} demonstrates impressive results on out-of-distribution dataset as well. The merit lies in its singular Fourier and inverse Fourier transformations, mapping input to output entirely in the frequency domain. 
    \item \textbf{Attention alone is not enough}: {\oformer}, employing the attention-based transformer architecture, showcases notable advantages on the {\navierstokes} dataset. It also demonstrates commendable results on specific PDEs like {\burgers}, {\darcy}. However, a glaring limitation surfaces when these architectures are applied to other PDEs, whether of comparable complexity or even simpler ones. They fail to generalize. This shortcoming starkly contrasts with one of the primary advantages anticipated from data-driven PDE solvers: the capacity to discern the solution operator solely from data, independent of prior knowledge of the underlying PDE.
    \item \textbf{Data-driven models work}: Surprisingly, the {\cgan}, a standard architecture for image tasks, excels in performance, even though it isn't inherently an operator. This prowess, however, wanes during cross-dataset evaluations, underscoring the importance of truly learning the underlying PDE rather than merely excelling on a given dataset.
    \item \textbf{Challenges with {\shear} and {\biaxial} Datasets}: The collective struggle of most operators with the {\shear} and {\biaxial} datasets underscores the importance of studying complex deformation patterns. Specifically, it suggests clear and well-defined failure modes in operators, where future works can be focused.
    \item \textbf{Time efficiency should be improved}: While the models give reasonable performance, they grapple with time efficiency. Especially, the best performing models such as transformer-based architectures are time-intensive both during training and inference, {\fno} is relatively swift in training, still intensive in inference. 
\end{enumerate}
{\bf Limitations and future work:} Although {\fno} and {\gnot} exhibit superior results, their inconsistent results in cross-dataset evaluations and zero-shot super-resolution raise the questions whether they are truly learning approximate solution to the underlying PDE. Similarly, although resilient to noise and OOD, the internal neural network architecture {\sno} remains largely unexplored and often yields subpar outcomes. Future endeavors leveraging {\sno} might pave the way to operators with improved robustness. Failure modes of operators in datasets require further investigations to build more robust operators that can capture complex shear deformations. Finally, the inference time of the model requires improvement so that they can be applied to largescale real-world problems.

\bibliographystyle{plain}
\bibliography{neurips_2023.bib}

\begin{thebibliography}{10}

\bibitem{bhattacharya2021model}
Kaushik Bhattacharya, Bamdad Hosseini, Nikola~B Kovachki, and Andrew~M Stuart.
\newblock Model reduction and neural networks for parametric pdes.
\newblock {\em The SMAI journal of computational mathematics}, 7:121--157, 2021.

\bibitem{brunton2022data}
Steven~L Brunton and J~Nathan Kutz.
\newblock {\em Data-driven science and engineering: Machine learning, dynamical systems, and control}.
\newblock Cambridge University Press, 2022.

\bibitem{cao2021choose}
Shuhao Cao.
\newblock Choose a transformer: Fourier or galerkin.
\newblock {\em Advances in neural information processing systems}, 34:24924--24940, 2021.

\bibitem{chen1995universal}
Tianping Chen and Hong Chen.
\newblock Universal approximation to nonlinear operators by neural networks with arbitrary activation functions and its application to dynamical systems.
\newblock {\em IEEE transactions on neural networks}, 6(4):911--917, 1995.

\bibitem{cybenko1989approximation}
George Cybenko.
\newblock Approximation by superpositions of a sigmoidal function.
\newblock {\em Mathematics of control, signals and systems}, 2(4):303--314, 1989.

\bibitem{debnath2005nonlinear}
Lokenath Debnath and Lokenath Debnath.
\newblock {\em Nonlinear partial differential equations for scientists and engineers}.
\newblock Springer, 2005.

\bibitem{fanaskov2022spectral}
Vladimir Fanaskov and Ivan Oseledets.
\newblock Spectral neural operators.
\newblock {\em arXiv preprint arXiv:2205.10573}, 2022.

\bibitem{hao2022physics}
Zhongkai Hao, Songming Liu, Yichi Zhang, Chengyang Ying, Yao Feng, Hang Su, and Jun Zhu.
\newblock Physics-informed machine learning: A survey on problems, methods and applications.
\newblock {\em arXiv preprint arXiv:2211.08064}, 2022.

\bibitem{hao2023gnot}
Zhongkai Hao, Zhengyi Wang, Hang Su, Chengyang Ying, Yinpeng Dong, Songming Liu, Ze~Cheng, Jian Song, and Jun Zhu.
\newblock Gnot: A general neural operator transformer for operator learning.
\newblock In {\em International Conference on Machine Learning}, pages 12556--12569. PMLR, 2023.

\bibitem{jagtap2021extended}
Ameya~D Jagtap and George~E Karniadakis.
\newblock Extended physics-informed neural networks (xpinns): A generalized space-time domain decomposition based deep learning framework for nonlinear partial differential equations.
\newblock In {\em AAAI spring symposium: MLPS}, volume~10, 2021.

\bibitem{jian2016deep}
S~Jian, H~Kaiming, R~Shaoqing, and Z~Xiangyu.
\newblock Deep residual learning for image recognition.
\newblock In {\em IEEE Conference on Computer Vision \& Pattern Recognition}, pages 770--778, 2016.

\bibitem{kingma2014adam}
Diederik~P Kingma and Jimmy Ba.
\newblock Adam: A method for stochastic optimization.
\newblock {\em arXiv preprint arXiv:1412.6980}, 2014.

\bibitem{kovachki2021neural}
Nikola Kovachki, Zongyi Li, Burigede Liu, Kamyar Azizzadenesheli, Kaushik Bhattacharya, Andrew Stuart, and Anima Anandkumar.
\newblock Neural operator: Learning maps between function spaces.
\newblock {\em arXiv preprint arXiv:2108.08481}, 2021.

\bibitem{lejeune2020mechanical}
Emma Lejeune.
\newblock Mechanical mnist: A benchmark dataset for mechanical metamodels.
\newblock {\em Extreme Mechanics Letters}, 36:100659, 2020.

\bibitem{li2022transformer}
Zijie Li, Kazem Meidani, and Amir~Barati Farimani.
\newblock Transformer for partial differential equations' operator learning.
\newblock {\em arXiv preprint arXiv:2205.13671}, 2022.

\bibitem{li2020fourier}
Zongyi Li, Nikola Kovachki, Kamyar Azizzadenesheli, Burigede Liu, Kaushik Bhattacharya, Andrew Stuart, and Anima Anandkumar.
\newblock Fourier neural operator for parametric partial differential equations.
\newblock {\em arXiv preprint arXiv:2010.08895}, 2020.

\bibitem{loshchilov2017decoupled}
Ilya Loshchilov and Frank Hutter.
\newblock Decoupled weight decay regularization.
\newblock {\em arXiv preprint arXiv:1711.05101}, 2017.

\bibitem{lu2021learning}
Lu~Lu, Pengzhan Jin, Guofei Pang, Zhongqiang Zhang, and George~Em Karniadakis.
\newblock Learning nonlinear operators via deeponet based on the universal approximation theorem of operators.
\newblock {\em Nature machine intelligence}, 3(3):218--229, 2021.

\bibitem{lu2022comprehensive}
Lu~Lu, Xuhui Meng, Shengze Cai, Zhiping Mao, Somdatta Goswami, Zhongqiang Zhang, and George~Em Karniadakis.
\newblock A comprehensive and fair comparison of two neural operators (with practical extensions) based on fair data.
\newblock {\em Computer Methods in Applied Mechanics and Engineering}, 393:114778, 2022.

\bibitem{mehran2022learning}
Meer Mehran~Rashid, Tanu Pittie, Souvik Chakraborty, and NM~Anoop~Krishnan.
\newblock Learning the stress-strain fields in digital composites using fourier neural operator.
\newblock {\em arXiv e-prints}, pages arXiv--2207, 2022.

\bibitem{mirza2014conditional}
Mehdi Mirza and Simon Osindero.
\newblock Conditional generative adversarial nets.
\newblock {\em arXiv preprint arXiv:1411.1784}, 2014.

\bibitem{nakamura1977computational}
Shoichiro Nakamura.
\newblock {\em Computational methods in engineering and science, with applications to fluid dynamics and nuclear systems}.
\newblock John Wiley and Sons, Inc., New York, 1977.

\bibitem{nelsen2021random}
Nicholas~H Nelsen and Andrew~M Stuart.
\newblock The random feature model for input-output maps between banach spaces.
\newblock {\em SIAM Journal on Scientific Computing}, 43(5):A3212--A3243, 2021.

\bibitem{raissi2019physics}
Maziar Raissi, Paris Perdikaris, and George~E Karniadakis.
\newblock Physics-informed neural networks: A deep learning framework for solving forward and inverse problems involving nonlinear partial differential equations.
\newblock {\em Journal of Computational physics}, 378:686--707, 2019.

\bibitem{robert2007partial}
D~Robert.
\newblock Partial differential equations and applications.
\newblock {\em S{\'e}min. Congr}, 15:181--250, 2007.

\bibitem{ronneberger2015u}
Olaf Ronneberger, Philipp Fischer, and Thomas Brox.
\newblock U-net: Convolutional networks for biomedical image segmentation.
\newblock In {\em Medical Image Computing and Computer-Assisted Intervention--MICCAI 2015: 18th International Conference, Munich, Germany, October 5-9, 2015, Proceedings, Part III 18}, pages 234--241. Springer, 2015.

\bibitem{sewell2012analysis}
Granville Sewell.
\newblock {\em Analysis of a finite element method: PDE/PROTRAN}.
\newblock Springer Science \& Business Media, 2012.

\bibitem{smith2019super}
Leslie~N Smith and Nicholay Topin.
\newblock Super-convergence: Very fast training of neural networks using large learning rates.
\newblock In {\em Artificial intelligence and machine learning for multi-domain operations applications}, volume 11006, pages 369--386. SPIE, 2019.

\bibitem{solin2005partial}
Pavel {\^S}ol{\'\i}n.
\newblock {\em Partial differential equations and the finite element method}.
\newblock John Wiley \& Sons, 2005.

\bibitem{takamoto2022pdebench}
Makoto Takamoto, Timothy Praditia, Raphael Leiteritz, Daniel MacKinlay, Francesco Alesiani, Dirk Pfl{\"u}ger, and Mathias Niepert.
\newblock Pdebench: An extensive benchmark for scientific machine learning.
\newblock {\em Advances in Neural Information Processing Systems}, 35:1596--1611, 2022.

\bibitem{targ2016resnet}
Sasha Targ, Diogo Almeida, and Kevin Lyman.
\newblock Resnet in resnet: Generalizing residual architectures.
\newblock {\em arXiv preprint arXiv:1603.08029}, 2016.

\bibitem{tripura2022wavelet}
Tapas Tripura and Souvik Chakraborty.
\newblock Wavelet neural operator: a neural operator for parametric partial differential equations.
\newblock {\em arXiv preprint arXiv:2205.02191}, 2022.

\bibitem{tristano1998advancing}
Joseph~R Tristano, Steven~J Owen, and Scott~A Canann.
\newblock Advancing front surface mesh generation in parametric space using a riemannian surface definition.
\newblock In {\em IMR}, pages 429--445, 1998.

\bibitem{vaswani2017attention}
Ashish Vaswani, Noam Shazeer, Niki Parmar, Jakob Uszkoreit, Llion Jones, Aidan~N Gomez, {\L}ukasz Kaiser, and Illia Polosukhin.
\newblock Attention is all you need.
\newblock {\em Advances in neural information processing systems}, 30, 2017.

\end{thebibliography}

\newpage
\appendix
\section{Appendix}
\subsection{Fluid Dynamics Datasets}\label{detailed description}
\subsubsection{Burgers Equation}
The 1D Burgers' equation is a non-linear PDE which is having the following form:
\begin{equation}
    \frac{\partial u}{\partial t}(x,t) + \frac{\partial}{\partial x}\left(\frac{u^2(x,t)}{2}\right) = \nu \frac{\partial^2 u}{\partial x^2}(x,t), \quad x \in (0,1), \, t \in (0,1]
\end{equation}
with periodic boundary conditions, where $u(x,0) = u_0(x)$, $x \in (0,1)$.

Here, $u_0 \in L^2_{\text{per}}((0,1);\mathbb{R})$ is the initial condition and $\nu \in \mathbb{R}^+$ is the viscosity coefficient.

\subsubsection{Darcy Flow Equation}
We experiment with the steady-state solution of 2D Darcy Flow over the unit square, whose viscosity term $a(x)$ is an input of the system. The following equation defines the solution of the steady state.
\begin{equation}
    \nabla \cdot (a(x) \nabla u(x)) = f(x), \quad x \in (0,1)^2, \label{eq:darcy}
\end{equation}
\begin{equation}
    u(x) = 0, \quad x \in \partial(0,1)^2. \label{eq:bc}
\end{equation}
For this paper, the force term $f$ is set to a constant value $\beta$, changing the scale of the solution $u(x)$. Instead of directly solving Equation \ref{eq:darcy}, we get the solution by solving a temporal evolution equation:
\begin{equation}
    \partial_t u(x,t) - \nabla \cdot (a(x) \nabla u(x,t)) = f(x), \quad x \in (0,1)^2, \label{eq:temporal}
\end{equation}

The equation is solved with random initial conditions until it reaches a steady-state solution.

\subsubsection{Compressible {\navierstokes} Equation}
We experimented with the 2-D Navier-Stokes equation for a viscous, incompressible fluid in vorticity form on the unit torus having the following form:

\begin{equation}
    \partial_t w(x,t) + u(x,t) \cdot \nabla w(x,t) = \nu \Delta w(x,t) + f(x), \quad x \in (0,1)^2, \, t \in (0,T] \\  
\end{equation}
\begin{equation}
        \nabla \cdot u(x,t) = 0, \quad x \in (0,1)^2, \, t \in [0,T] \\
\end{equation}
\begin{equation}
        w(x,0) = w_0(x), \quad x \in (0,1)^2
\end{equation}

where $u \in C([0,T]; H^r_{\text{per}}((0,1)^2; \mathbb{R}^2))$ for any $r > 0$ is the velocity field, $w = \nabla \times u$ is the vorticity, $w_0 \in L^2_{\text{per}}((0,1)^2; \mathbb{R})$ is the initial vorticity, $\nu \in \mathbb{R}^+$ is the viscosity coefficient, and $f \in L^2_{\text{per}}((0,1)^2; \mathbb{R})$ is the forcing function.

\subsubsection{Incompressible {\navierstokes} Equation}
The incompressible Navier-Stokes equations represent a specialized form of the broader compressible fluid dynamics equations, tailored for scenarios involving subsonic flows. These equations are versatile and can be employed to analyze a wide range of systems, from hydromechanical processes to meteorological predictions and the exploration of turbulent behaviors.

\subsubsection{Shallow Water Equation}
The shallow-water equations present a suitable framework for modeling free-surface flow problems. Its 2D hyperbolic PDEs with the following form:
\begin{equation}
    \partial_t h + \nabla \cdot (hu) = 0, \\
\end{equation}

\begin{equation}
    \partial_t hu + \nabla \cdot \left(u^2h + \frac{1}{2}grh^2\right) = -grh\nabla b, \label{eq:momentum}
\end{equation}
where $u$ and $v$ are the velocities in the horizontal and vertical directions, $h$ denotes the water depth, and $b$ describes spatially varying bathymetry. $hu$ represent as the directional momentum components and $g$ represent the gravitational acceleration. 

\subsection{Mechanical Datasets} \label{mechanical}

\subsubsection{Stress, Strains fields in 2D Digital in Composites}
The constitutive relationship (generalized Hook's law)  is defined as :
\begin{equation}
    \{\sigma_{ij} \} = \mathbb{C}_{ijkl} \{\epsilon_{kl}\}
\end{equation}
where $\sigma_{ij}$ and $\epsilon_{kl}$ are the stress and strain components, $\mathbb{C}_{ijkl}$ is the overall stiffness tensor.
For 2D problems, we have $\sigma_{xx}$ , $\sigma_{yy}$ and $\sigma_{xy}$ satisfy the following equilibrium equations
\begin{equation}
    \partial_x \sigma_{xx} + \partial_y \sigma_{xy} + F_{x} =0 ; 
    \partial_y \sigma_{yy} + \partial_x \sigma_{xy} + F_{y} =0  
\end{equation}
where $F_x$ and $F_y$ are the body forces in horizontal and vertical directions, respectively. Additionally, the strains $\epsilon_{xx}$, $\epsilon_{yy}$ and $\epsilon_{xy}$ are defined as 
\begin{equation}
    \epsilon_{xx} = \partial_x u_{xx} ; \epsilon_{yy} = \partial_{y} u_{yy} ;\epsilon_{xy}= \partial_y u_x + \partial_x u_x
\end{equation}
where $u_x$ and $u_y$ are the displacements in horizontal and vertical directions, respectively
\\ The 2D-Digital composites\citep{mehran2022learning} are subjected to mode-I tensile loading with specific boundary conditions. The two-phase composite has a modulus ratio of 10 $(E_{stiff}/E_{soft})$, and both materials are assumed to be perfectly elastic. The simulations are run on an $8mm \times 8mm$ plate with negligible thickness. The two phases are randomly placed over the plate; however, the fraction of soft and stiff phases is equal.

\subsubsection{Mechanical MNIST}
The mechanical MNIST \citep{lejeune2020mechanical} has a collection of different tests such as shear, equibiaxial extension, confined compression, and other tests simulated on a heterogenous material subjected to large deformations. The famous MNIST digit dataset is generated by treating the bitmap images as heterogeneous material blocks modeled as a Neo-Hokean material of varying modulus. The material is subjected to fixed displacements in one direction, and the full-field displacements are recorded at each step. In this study, we select two tests from the mechanical MNIST experiments viz shear and equibiaxial. 

\end{document}